\documentclass{article}

\usepackage[final]{corl_2019} 

\usepackage{amsmath} 
\usepackage{amssymb}  
\usepackage{color}
\usepackage{hyperref}

\usepackage[bottom]{footmisc}

\usepackage{graphicx} 
\usepackage{caption}
\usepackage{subcaption}

\usepackage{gensymb} 

\usepackage{enumerate} 
\usepackage{enumitem}

\usepackage{soul}

\usepackage{wrapfig}
\usepackage[textsize=scriptsize,textwidth=3.cm]{todonotes}

\renewenvironment{quote}{%
   \list{}{%
     \leftmargin0.5cm   
     \rightmargin\leftmargin
   }
   \item\relax
}
{\endlist}

\newcommand\blfootnote[1]{%
  \begingroup
  \renewcommand\thefootnote{}\footnote{#1}%
  \addtocounter{footnote}{-1}%
  \endgroup
}

\title{Task-Informed Fidelity Management for \\ Speeding Up Robotics Simulation}

\author{
  Abhijeet Tallavajhula$^*$ \\
  National Robotics Engineering Center \\
  Carnegie Mellon University \\
  \texttt{abhijeet.t91@gmail.com} \\
  \And
  Adrian Schoisengeier$^*$ \\
  National Robotics Engineering Center \\
  Carnegie Mellon University \\
  \texttt{aschoise@andrew.cmu.edu} \\
  \And
  Sung-Kyun Kim \\
  Jet Propulsion Laboratory \\
  California Institute of Technology \\
  \texttt{sung.kim@jpl.nasa.gov} \\
  \And
  Anirudh Vemula \\
  Robotics Institute \\
  Carnegie Mellon University \\
  \texttt{avemula1@andrew.cmu.edu} \\
  \And
  Levi Lister \\
  National Robotics Engineering Center \\
  Carnegie Mellon University \\
  \texttt{levi.lister@gmail.com} \\
  \And
  Oren Salzman \\
  Computer Science Department \\
  Technion-Israel Institute of Technology \\
  \texttt{salzman.oren@gmail.com} \\
}

\begin{document}
\maketitle


\begin{abstract}
  Simulators are an important tool in robotics that is used to develop robot software
  and generate synthetic data for machine learning algorithms.
  Faster
  simulation can result in better software validation and larger amounts of
  data.
  Previous efforts for speeding up simulators have been performed at the level
  of simulator building blocks, and robot systems.
  Our key insight, motivating
  this work, is that further speedups can be obtained at the level of the
  \emph{robot task}.
  Building on the observation that not all parts of a scene
  need to be simulated in high fidelity at all times, our approach is to toggle
  between high- and low-fidelity states for scene objects in a task-informed
  manner.
  Our contribution is a framework for speeding up
  robot simulation by exploiting task knowledge.
  The framework is agnostic to the
  underlying simulator, and preserves simulation fidelity.
  As a case study, we consider a complex material-handling task.
  For the
  associated simulation, which contains many of the characteristics that make
  robot simulation slow, we achieve a speedup that can be up to three times faster than high fidelity without compromising on the quality of the results.
  We also
  demonstrate that faster simulation allows us to train better policies for
  performing the task at hand in a short period of time.
  A video summarizing our contributions can be found at \url{https://youtu.be/PEzypDyqc3o}.
\end{abstract}



\section{Introduction} \label{sec:introduction}

Robot\blfootnote{$^*$ 
A. Tallavajhula and 
A. Schoisengeier
contributed equally to this work.} systems and tasks are becomingly increasingly complex as they progress
from labs to real applications. Developing and testing robot software can be
challenging in the real world and leveraging simulators is one solution to these
challenges~\cite{aguero2015inside}, \cite{vzlajpah2008simulation}. Simulators
are also promising as a source of data for machine learning algorithms which are
often used in robotics~\cite{tobin2017domain}, \cite{ros2016synthia}. In
simulation, scenes can be setup according to task and algorithm specification,
and complete state information is readily available---a critical requirement for
many algorithms.  While \emph{fidelity} or accuracy is arguably the most
important aspect of simulators, another critical feature is \emph{speed}. Based
on a user survey~\cite{ivaldi2014tools}, speed was determined to be the
third-most important criterion for selecting a robot simulator, after fidelity,
and access to open-source software. There is an inherent trade-off between
simulator fidelity and simulator speed~\cite{erez2015simulation}, so that
running simulation at the highest fidelity settings may slow it down to the
extent where speed becomes the computational bottleneck. Slow simulation can
mean incomplete validation of robot software, or insufficient volumes of
training data, motivating efforts to speed up simulation.

Past efforts on speeding up simulation have been focused at different levels,
starting with the fundamental \emph{building blocks} of robot simulators such as
the physics engines, object rendering, and more. These include straightforward
optimizations such as simplifying object meshes~\cite{thakur2009survey}, to
advanced improvements such as advanced dynamics
solvers~\cite{zapolsky2015adaptive}. Simulation speedups have also been
performed at the level of \emph{robot systems} where domain knowledge can be
incorporated.  Examples include platform-specific dynamics
updates~\cite{hereid2017frost}, and subdivision of simulation
scenes~\cite{pinciroli2011argos}. Our key insight is that there is further
potential for speedup at a higher level, namely the \emph{robotics task} of
interest. Speedups obtained at building-block and robot-system levels are often
general, and may not be the appropriate place to express domain knowledge. Our
work introduces a formulation for exploiting task-level knowledge for simulation
speedup.

\begin{figure}[t]
%
 \begin{subfigure}{0.32\textwidth}
  \centering
\includegraphics[trim=200 150 550 300,clip, width=0.99\textwidth]{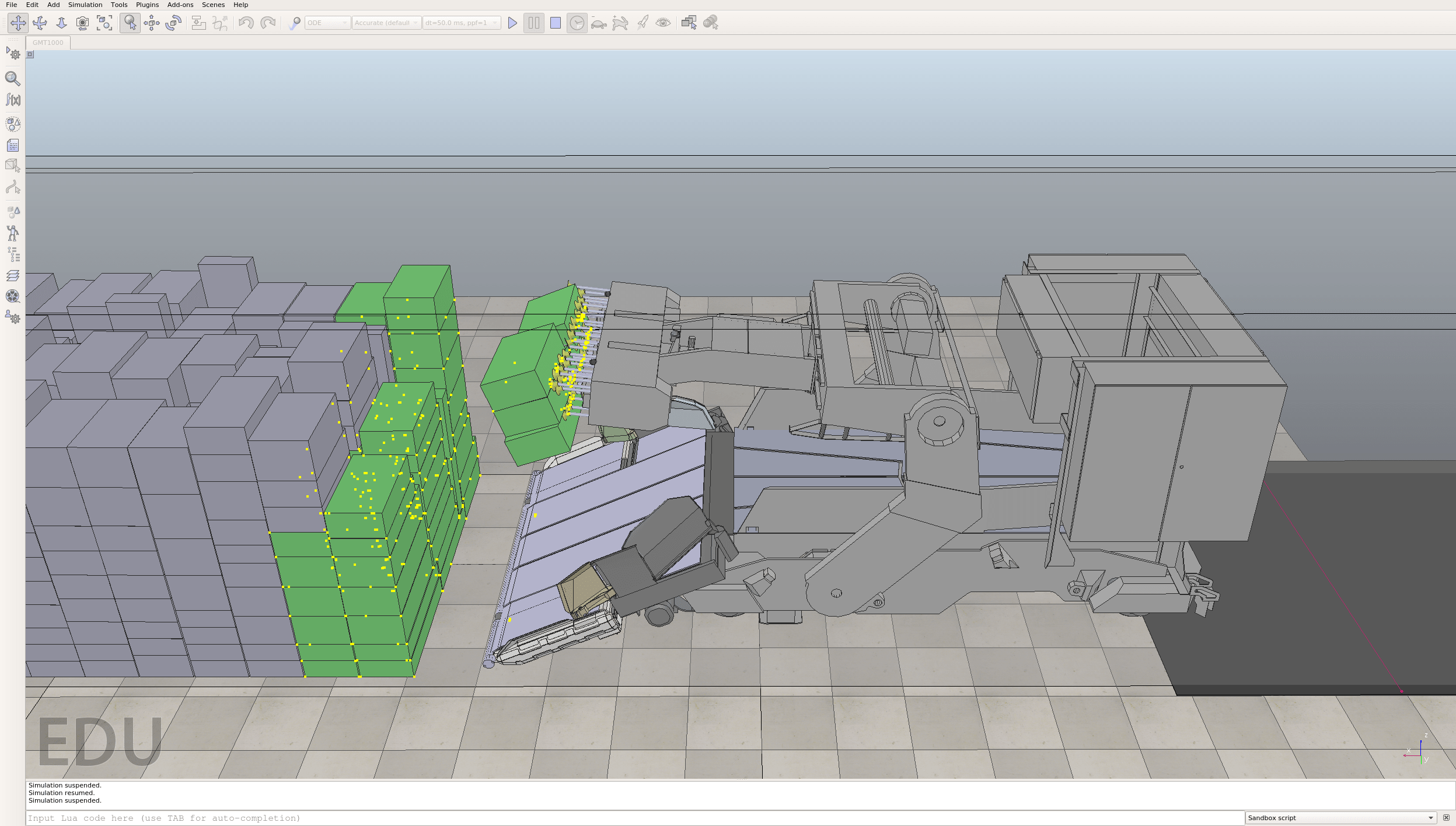}
  \caption{}
  \label{fig:intro-pick}
 \end{subfigure}	
 \begin{subfigure}{0.32\textwidth}
  \centering
\includegraphics[trim=200 150 550 300,clip, width=0.99\textwidth]{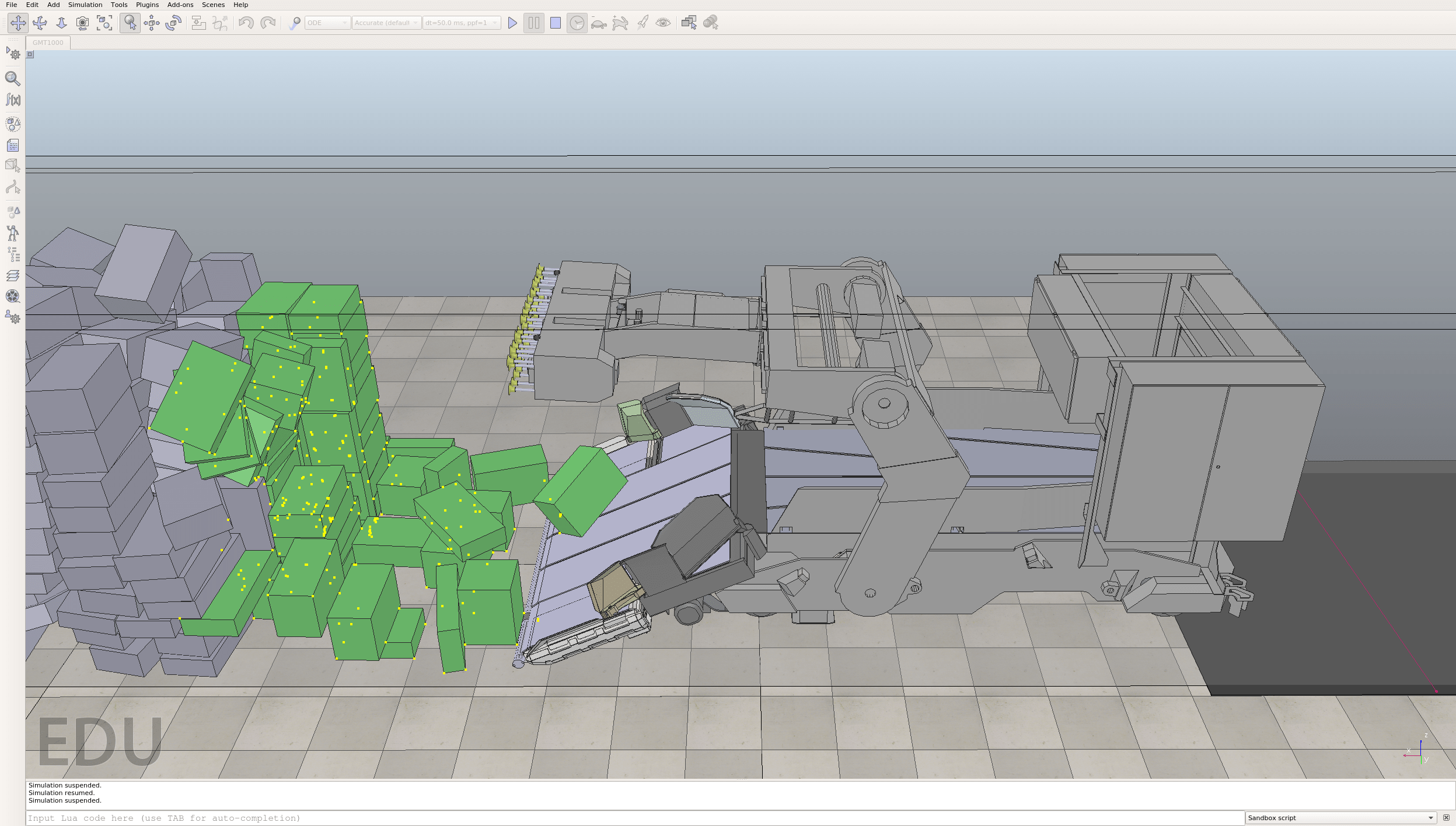}
  \caption{}
  \label{fig:intro-sweep}
 \end{subfigure}	
 \begin{subfigure}{0.32\textwidth}
  \centering
\includegraphics[trim=200 150 550 300,clip, width=0.99\textwidth]{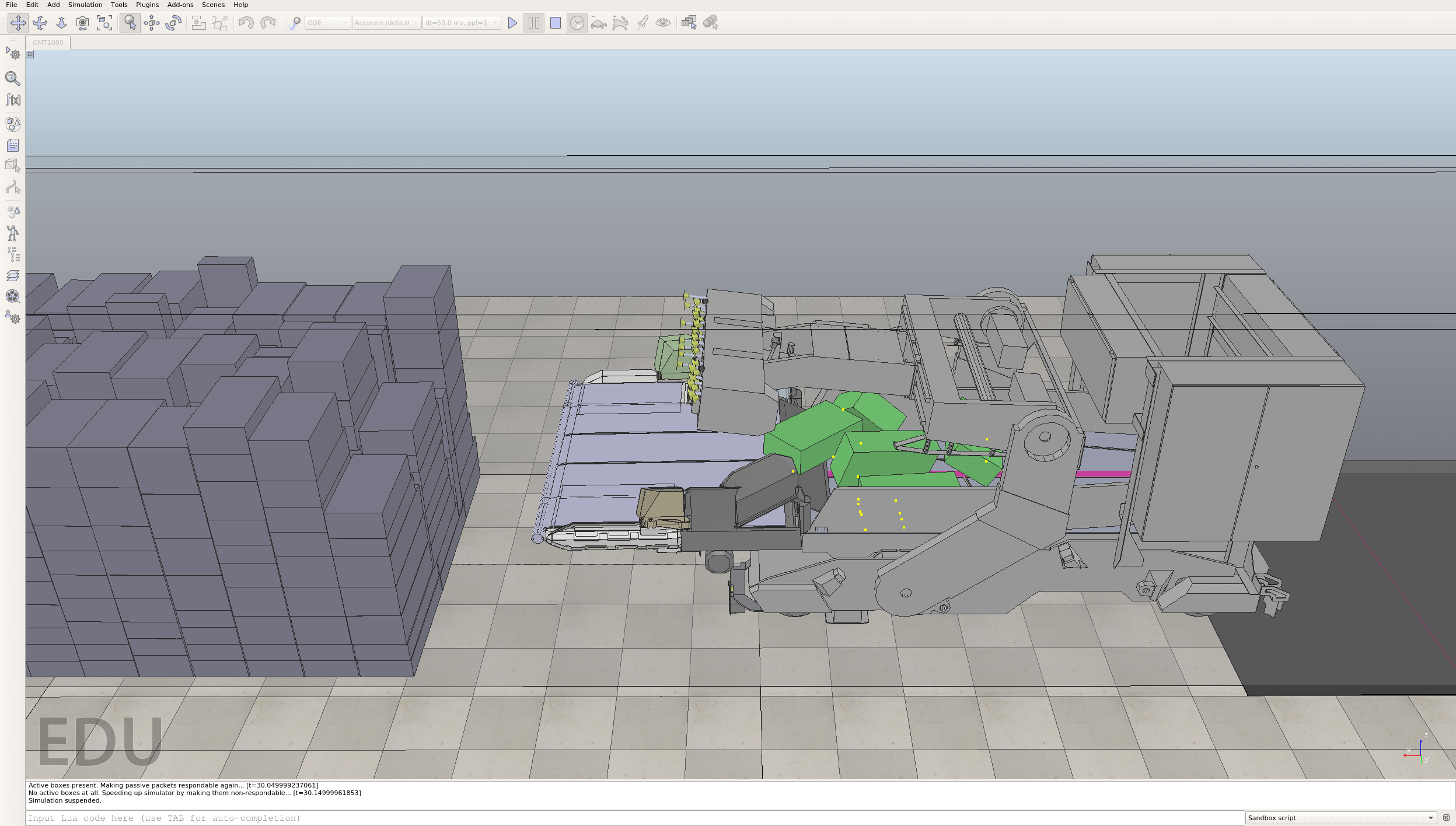}
  \caption{}
  \label{fig:intro-discharge}
 \end{subfigure}	
  \caption{Illustration of our approach to task-informed fidelity management for a material-handling application of truck unloading.
  The robot (see also Sec.~\ref{sec:implementation} and Fig.~\ref{fig:gmt}) can pick boxes using either 
  (\subref{fig:intro-pick})~a custom-designed end effector  
  or
  (\subref{fig:intro-sweep})~a scooper-like tool. 
  (\subref{fig:intro-discharge})~Once product is picked, it is discharged through the robot onto conveyor belts.
  Using our approach, only product that is relevant to the task at hand is simulated in high fidelity (green boxes with yellow contact points). The rest of the product is simulated in low fidelity resulting in a speedup in the simulation. }
  \label{fig:approach}
\end{figure}

Our approach, depicted in Fig.~\ref{fig:approach}, is based on the observation that not all objects in a simulation
scene need to be simulated in high fidelity at all times. We define high- and
low-fidelity states for different objects. The logic of managing fidelity
resides within the scene. As a side consequence, application-level robot
software is agnostic to the simulator, and can be ported directly to
hardware. Our framework can theoretically be applied to any general-purpose
robotics simulator, and can provide speedups alongside independent improvements
such as parallelization. As a case study, we consider a specific application
related to robotic material handling. The task involves a full-scale robot in an
industrial scenario where we found that naively using the simulator  proved to be too slow to be useful.

The rest of this paper is organized as follows. Related work is discussed in
Sec.~\ref{sec:related_work}. Our approach for fidelity management is
formulated in Sec.~\ref{sec:approach}. Details of the task and associated
simulation of our case study are in Sec.~\ref{sec:implementation}. In Sec.~\ref{sec:results} we present results for the simulation speedup, impact on task
performance, and generality of our approach. We conclude in Sec.~\ref{sec:conclusion}, and mention avenues for future research.

\section{Related work} \label{sec:related_work}
A robot simulator consists of components related to dynamics, sensing, and
visualization. These building blocks are obvious targets for optimizations and
speedups. 
For example, Drumwright et al.~\cite{drumwright2010extending} reformulated the dynamics update equations of a
physics engine (ODE~\cite{smith2005open}), and solved them in parallel. 
Other examples include changing the integration steps used by solvers---while
dynamics solvers often use fixed, first-order integration steps, Zapolsky and
Drumwright~\cite{zapolsky2015adaptive} proposed an algorithm with adaptive step sizes, which could lead to faster
dynamics simulation.  
The physics engine introduced by Thulesen et al.~\cite{thulesen2016robworkphysicsengine} used second-order integration for
greater accuracy. In addition, by modeling contact handling as a convex-optimization problem, a speedup was obtained over classic dynamics solvers,
which use a Linear-Complementary Problem formulation. 
Wakisaka et al.~\cite{wakisaka2016fast} considered dynamics at the motor-joint level, and a
fast method to compute forward dynamics for robot manipulators was proposed. 
Finally, Crozet et al.~\cite{crozet2016fast} considered a fine-grained sub-problem of
computing contact points to speed up simulation. 

Other building blocks that can be altered to speed up robotic simulators are
rendering and visualization.
OptiX~\cite{parker2010optix} is an example of parallelizing ray-tracing, an
operation also used to simulate sensors.
For fast visualization, Awaad and Le{\'o}n~\cite{awaad2008xpersim} propose a
setup of a single simulation server, and multiple clients. Speedup was achieved
through functional separation, by performing dynamics updates on the
server-side, and visualization on the client-side.


While improvements to the fundamental building blocks of simulators are generic, further speedups may be obtained by exploiting characteristics of the robot systems under
consideration. 
For example, Huerzeler et al.~\cite{huerzeler2013configurable} introduced a simulator for coaxial
rotor unmanned aerial vehicles (UAVs), for which dynamics equations specific to the platforms were derived from first principles. The resulting simulation structure and equations were compiled into binaries using SIMULINK\footnote{\url{https://www.mathworks.com/products/simulink.html}}, which resulted in real-time simulation. 
A simulator for UAVs was also presented
by Symington et al.~\cite{symington2014simulating}. While simulation speed was not the focus,
domain knowledge was used in the modeling stage to balance fidelity with
real-time performance. 
%
A similar approach was taken for legged locomotion~\cite{hereid2017frost}: a dynamics model for the systems of interest was developed, and
an optimized representation of symbolic equations was deployed to improve simulation
speed.  
In the domain of swarm robots, simulation speed is not just a convenience, but a bottleneck. 
Pinciroli et al.~\cite{pinciroli2011argos}, developed a semaphore-free multi-threaded implementation of the main simulation
loop. Furthermore, the simulation scene was partitioned, and each sub-space allocated to a separate physics engine. With these optimizations, $10, 000$ robots could be simulated faster than real-time. 
The scale of robot modules being simulated was pushed to a million~\cite{ashley2011simulating}. Here, various parts of the physics engine
(collision detection, dynamics solving) were parallelized via a deep familiarity
with the underlying computer architecture. In addition, simulation for modules
was distributed over a cluster. 
Speedups for the modular robot simulator
presented by Collins et al.~\cite{collins2013remod3d} were primarily achieved by relying on the
parallel processing ability of the
PhysX\footnote{\url{https://www.geforce.com/hardware/technology/physx}} physics
engine. GPU-based dynamics calculation of hydraulic components was also used for
simulating soft robotics~\cite{rodriguez2017real}.

Related to our material handling task, the work by Miklic et
al.~\cite{miklic2012modular} used USARSim~\cite{carpin2007usarsim} to simulate
the transportation of goods by mobile robots in a warehouse. Specifically for a
scheduling algorithm, a bare-bones simulator was used. Their work can be seen as
a simple case of using task-level information to design a fast, low-fidelity
simulator. While speed may not be the main focus of past work, results
presented were often for isolated joints or a few interacting
bodies~\cite{thulesen2016robworkphysicsengine},~\cite{wakisaka2016fast}.
Similar to recent work~\cite{echeverria2011modular}, we stress the importance of performing experiments
for an entire robot system for evaluating simulators.

Simulations play a crucial role in learning applications where they
are often used as a forward model to plan optimal policies on. Recent
work \cite{mordatch2015ensemble, peng2018sim, hwangbo2019learning,
  tan2018sim} have used simulation effectively to learn control
policies for a wide variety of robots ranging from quadrupeds and
legged robots to humanoid robots. These approaches demand fast
simulation speeds as they need a large amount of data to be collected
to fit approximations of forward models or directly learn control policies. However,
if the fidelity of the simulation decreases, the learned policy will
also suffer in performance. Hence, there is a need for both high fidelity
and fast simulation speed in robotic simulations for control
applications. To achieve this, previous work has either introduced
noise into the low-fidelity simulations so that resulting policies are
robust to noise \cite{peng2018sim, tobin2017domain, vuong2019pick}, or
approximate inaccurate parts of the simulation with fast inference
models such as neural networks \cite{hwangbo2019learning} which are
learned using real-world data. Our proposed approach differs by
directly managing the fidelity of the simulation 
based on the inherent structure of the robot task.

\section{Approach} \label{sec:approach}

Since high-fidelity simulation can be time consuming, our idea is to actively
toggle parts of a scene to a low-fidelity mode, when they are not needed to be
simulated accurately. Our aim is to formulate a principled approach for
task-informed fidelity management. In our view, any such approach should satisfy
the following requirements.
\begin{enumerate}[label={\textbf{R\arabic*}}]
\item \label{req:1} It should minimize the loss of simulation fidelity incurred due to increase in speed.
\item \label{req:2} It should not be computationally intensive. If this were the case, the speedup gain would not be worth the loss due to the management itself.
\item \label{req:3} It should be agnostic to the underlying layers of the physics engine, renderers, and other simulation building blocks.
\item \label{req:4} It should, in turn, appear transparent to any robot-application software (such as planning, and perception). It is good practice to develop applications that are agnostic to simulation; this requirement is simply an extension of the practice.
\item \label{req:5} It should exhibit good software features, such as modularity, and reusability. If the approach was cumbersome to implement and use, configuring it may not be worth a developer's time.
\end{enumerate}

\begin{figure}[tb]
  \centering
  \begin{subfigure}{0.49\textwidth}
    \includegraphics[width=0.9\textwidth]{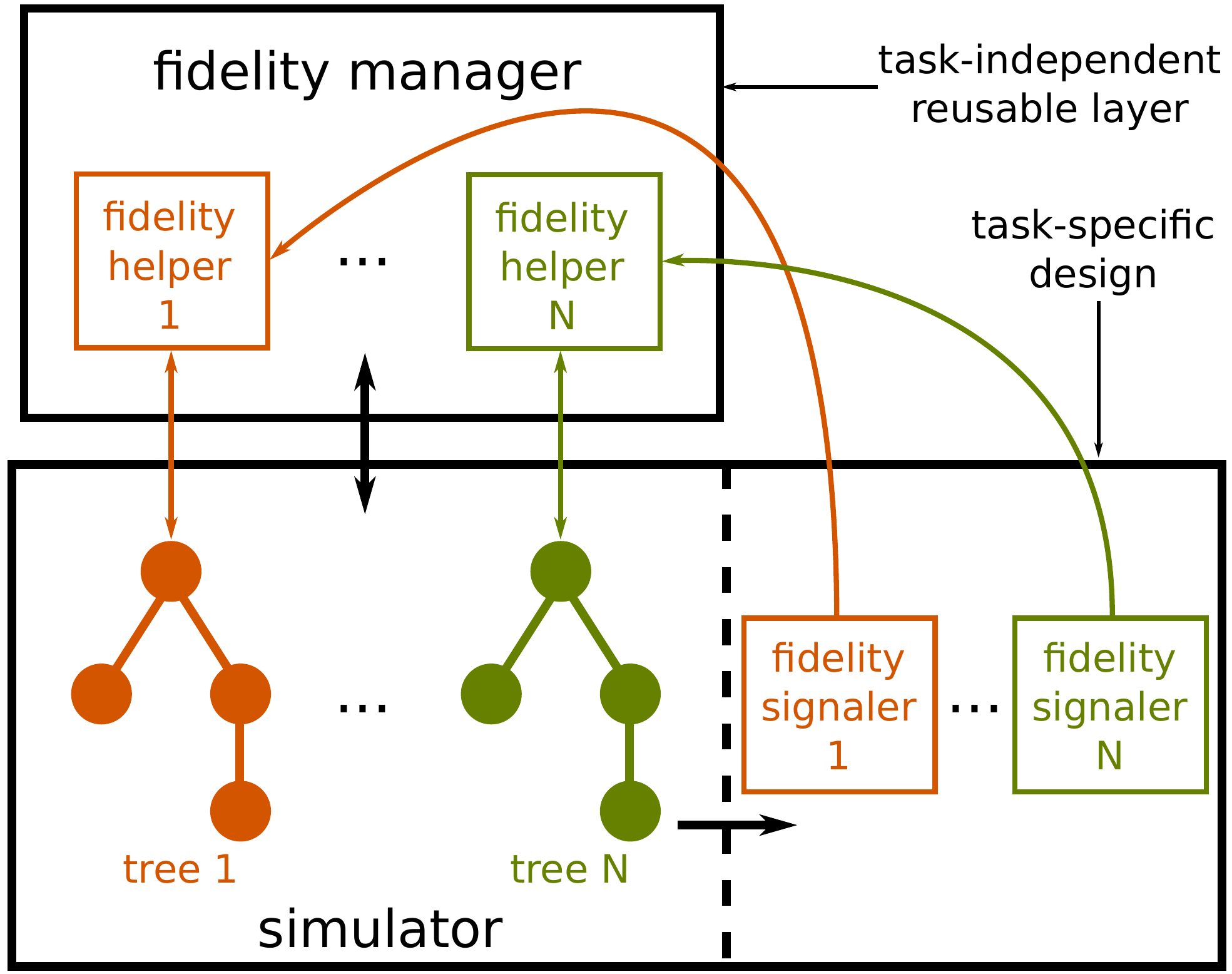}
    \caption{Flow of information between the simulator, fidelity helpers,
      and signalers.}
    \label{fig:architecture}
  \end{subfigure}
  \hfill
  \begin{subfigure}{0.49\textwidth}
    \includegraphics[width=0.9\textwidth]{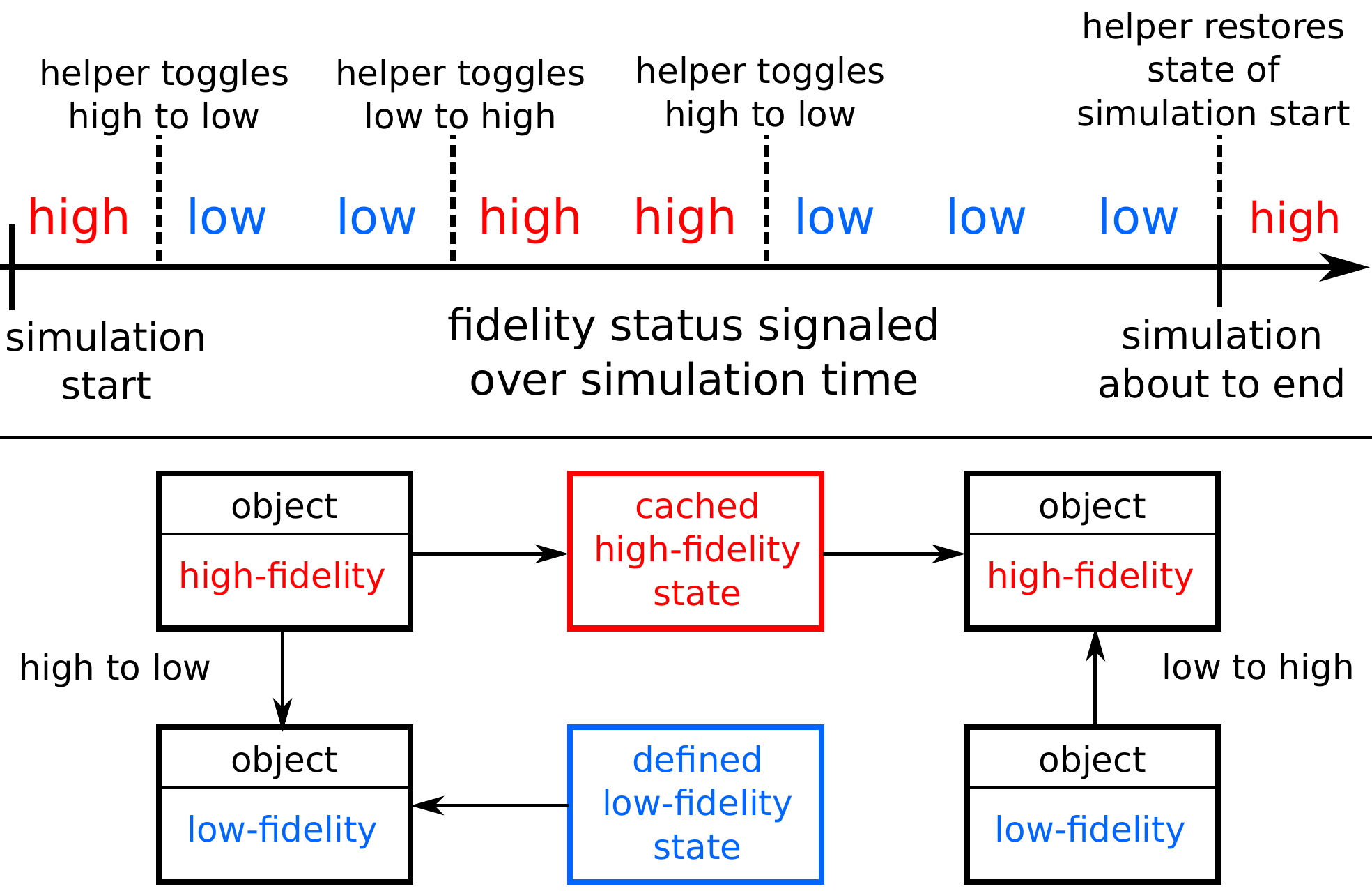}
    \caption{Example of fidelity toggling in response to received fidelity
      signals. Below, depiction of the cached high-fidelity state, and common
      low-fidelity state.}
    \label{fig:helper_logic}
  \end{subfigure}
  \caption{Fidelity manager architecture}
\end{figure}
  
Our proposed architecture is depicted in Fig.~\ref{fig:architecture}. An
\emph{object} is any entity in the simulation scene whose fidelity can be
toggled. Objects can include shapes with masses, and sensors. A \emph{tree} is a
collection of objects, organized in a tree data structure. For example, a robot
is commonly expressed in simulation as a tree. The fidelity \emph{manager}
consists of a number of fidelity \emph{helpers}. Each helper is associated with
a tree in a simulation scene, and is responsible for toggling its fidelity. Each
tree is associated with a fidelity \emph{signaler}, which communicates the
desired current fidelity state (high or low) to the corresponding helper. The
signaler's computation is ideally lightweight, to satisfy~\ref{req:2}. Note that all objects of a particular type are treated
identically, while each tree is handled independently.

The operation of each fidelity helper is depicted in Fig.~\ref{fig:helper_logic}. A tree's fidelity is toggled only when there is a change
in the fidelity status, as reported by the signaler. For every object type, a
common low-fidelity state is defined. When toggling an object from high to low
fidelity, the high-fidelity state is cached, and the common low-fidelity state
is applied. 
When
toggling back from low to high fidelity, the cached fidelity state is
applied. When simulation ends, the helper resets the fidelity state of the tree
to that of when simulation was started. With these choices, intricate simulation
settings of objects are preserved, and the manager leaves no side effect in the
scene.

We have separated deciding what the fidelity state of a tree should be
(performed by the fidelity signaler) from the actual toggling (performed the
fidelity helper). The signalers are more closely connected to the simulation
scene than the manager, depicted by a dashed separating line in Fig.~\ref{fig:architecture}. As a consequence of the separation, the manager is a
software layer that can be reused across tasks. For a new scene, a developer
only needs to encode domain knowledge in the signalers. These design choices
were made keeping~\ref{req:5} in mind.


As for the remaining requirements, there is no flow of information from the
fidelity manager or signalers to robot applications, satisfying~\ref{req:4}. In our approach, the fidelity states only access the simulator's
API, not the underlying physics engine, in agreement with~\ref{req:3}. 
Our framework can be combined
with other methods for speedup, such as parallelization. While not the scope of
this work, we note that parallelizing over trees is a possible implementation,
as they are independent. Requirement~\ref{req:1}, that the increased speed does not come at the cost of low simulation fidelity, is validated empirically in Sec~\ref{sec:results}.

\section{Implementation} \label{sec:implementation}
In this section, we turn to a concrete robot task, describe the simulation
setup, and instantiate the architecture described in Sec.~\ref{sec:approach}.
These form the basis for the experiments and results described in Sec.~\ref{sec:results}.

\subsection{Task---truck unloading for warehouses}


\begin{wrapfigure}{r}{0.4\textwidth}
  \begin{center}
  \includegraphics[trim=200 0 200 0,clip, width=0.4\textwidth]{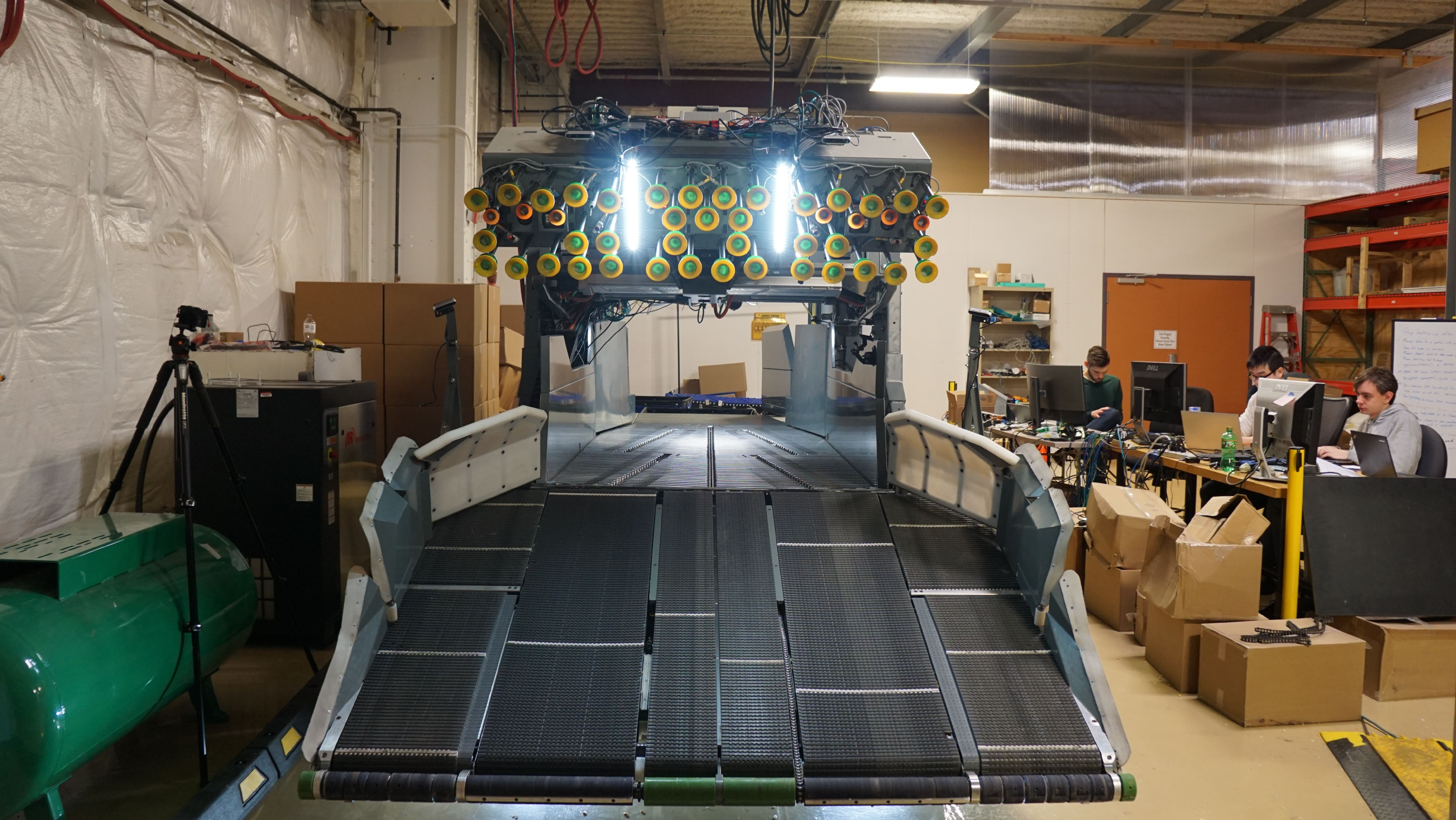}
  \end{center}

  \caption{Truck-Unloading Robot with manipulator-like tool with suction cups referred to as ``arm'' and a scooper-like tool with conveyor belts referred to as ``nose''.}
  \label{fig:gmt}
\end{wrapfigure}
Our task is one of robotic material handling where we are required to unload
packets from a truck onto a warehouse conveyor belt using a custom-designed
robot. Shipment trucks which transport goods to a warehouse are currently
unloaded by manual labor, which is time-consuming, labor intensive and
expensive. There is increasing interest in automating this and similar domains,
and the material-handling robotics market is projected to reach a size of $\$31$
billion by
$2023$\footnote{\url{https://www.marketresearchfuture.com/reports/material-handling-robotics-market-5099}}.
Our task is thus motivated by real-life applications, with the potential for
large-scale impact.

The mobile truck-unloading robot we work with, shown in Fig.~\ref{fig:gmt},
consists of two end effectors---a manipulator-like tool referred to as the
``arm'' and a scooper-like tool referred to as the ``nose''. The arm's
function is to \emph{pick} stacked packets, as it has a tool with a number of
suction cups mounted on coaxial plungers, while the nose's function is to
\emph{sweep} up packets near the floor. Conveyor belts, along the nose and body
of the robot, transfer packets to the back of the robot and onto the warehouse
conveyor belt. Packets which are transferred, via robot actions, from the truck
to the back of the robot, are considered as unloaded. The task is to unload a
full truckload, while maximizing the unloading rate, and the fraction of packets
unloaded.

Given the robot size and problem complexity, an accurate and fast simulation is
indispensable to solve this task. It is used to test the software stack and
optimize strategies to unload the robot prior to their execution on the real
machine. As the state of all packets is known, the simulation allows to compute
various performance metrics and compare different approaches. It also allows to
reduce development time as delays due to hardware and perception issues can be
mitigated by simulation. Furthermore, damage to the robot is prohibitively
expensive, and the safety of actions can be tested in simulation prior to
reality. Finally, setting up test scenarios, tedious in the real world, is
simple in simulation.

\subsection{Simulation setup}
For our task, we used the open-source, cross-platform simulation tool
V-REP~\cite{rohmer2013v}. Two main requirements led us to this decision. First,
given the complexity of the truck-unloading system, a high-accuracy simulator
with tunable physics parameters is needed to achieve sufficient correspondence
between reality and simulation~\cite{jakobi1995noise}. Second, given our
objective to dynamically toggle between fidelity levels, the simulator should
offer multiple options to programmatically customize the scene at runtime. The
simulator comparison~\cite{ivaldi2014tools} highlighted the strength of V-REP in
these two regards. Moreover, V-REP was one of the highest-ranked platforms in the 
aforementioned user study.

In the first stage of this project, we created a high-fidelity representation of
our truck-unloading system in V-REP. The robot was modeled at full scale with 10
active joints for the arm and nose, and 180 passive joints to model the 45
flexible plungers. By testing three of the built-in physical engines (namely, 
Bullet, ODE, and Vortex) on a set of representative packet pick-and-place 
scenarios in V-REP, we concluded that ODE provides the most suitable balance 
between accuracy and speed for our task.

In addition, several experiments were conducted on both the real and simulated
robot to tune simulation parameters, such as the friction properties of packets
and conveyor belts, the PID control values of active joints, and the
spring-damper values of passive joints. 
During these experiments, we used packet dimensions and weights that are typical 
in the shipping industry.

\subsection{Simulation fidelity management} \label{ssec:simulation-fidelity-management}

As a result of the accurate modeling of the real system, our simulation
contained many components that makes robot simulation computationally
expensive. Simulating the packets in the truck required multibody dynamics. The
numerous physical interactions (arm-packet, nose-packet, conveyor-packet,
packet-packet), required elaborate rigid-body dynamics for handling impacts,
contacts, and friction. Finally, as part of the system-level simulation,
numerous \emph{scripts} were included. A script can be thought of as a local
unit of computation which updates a part of the simulation scene. While each
script is lightweight, many scripts together can slow down simulation.

To adapt the fidelity of the different simulation components, we used the
fidelity manager described in Sec.~\ref{sec:approach}. The object types
chosen were \emph{shapes, joints, sensors}, and \emph{scripts}. We defined low
fidelity states for each object type as follows: low-fidelity shapes were static
and non-respondable; joints were passive; sensors needed explicit handling; and
scripts were disabled. As the signaling mechanism, we used V-REP signals.
Profiling revealed that the most computationally expensive components to 
simulate were the plungers in the robot arm, and packets in the truck. We 
describe implementation details for each separately.

\textbf{Plungers in arm}. The robot arm contains 45 plungers, each of which was
modeled using four links, four joints, a proximity sensor, and a
script. We assigned a fidelity helper for each plunger, treating it as a
tree. The task-specific information we implemented was that the plungers could
be simulated in low-fidelity when the robot was not executing a pick
action. Right before a pick (namely, as the arm approaches the boxes and interaction between the two is expected to occur), the planner commands the plungers to extend. The
signaler would detect this, and set the fidelity status to high. After a pick,
when the planner commanded the plungers to retract (and no further interaction is expected to occur), the signaler would set the
fidelity status to low. This design of the signaler ensured that the fidelity
state of the arm was consistent with what the task planner expected.

\textbf{Packets in truck}. A typical truck load contains several hundreds, and sometimes thousands of packets. 
Since the robot can only access packets in the
front of the truck, our task-specific implementation involves simulating packets
located in the back of the truck with a lower fidelity state. We created a
simulation script that treats each packet in the scene as a tree with one
object. The fidelity helper switches between three possible states. Packets in
high-fidelity are dynamic and respondable, packets in medium-fidelity are
non-dynamic and respondable, and packets in low-fidelity are non-dynamic and
non-respondable.

The signaler tracks the position of each packet relative to an inflated bounding box of the robot. Packets located inside this volume are set to high-fidelity, whereas packets entirely outside the volume are set to medium-fidelity. 
In situations where the robot cannot physically interact with
any packets in the truck, the signaler lowers the state of all
medium-fidelity packets to low-fidelity. 
Note that packets in low-fidelity cannot physically interact with any dynamically-enabled shapes (i.e. packets in high- or medium-fidelity). 
Therefore, the switch from medium- to low-fidelity is
only commanded when the signaler detects that all high-fidelity packets are entirely within the robot bounding-box borders. 
This guarantees that there is no physical contact between any dynamically-enabled objects inside the bounding box and any objects outside the bounding box.

As mentioned above, a low- or medium-fidelity packet that partially enters the robot bounding box will be elevated to high-fidelity immediately by the signaler. 
However, for a packet to be lowered from high-fidelity to a lower fidelity state, it would have to be entirely outside the bounding box and not have any linear or angular velocity. 
Once these two conditions are met for the period of two simulation steps, the signaler lowers the packet's fidelity state. Waiting for two simulation steps compensates for possible errors of the simulator, which might cause a very slowly moving object to appear still for one simulation step. 

In addition to the above rules, the signaler sets all packets to high-fidelity at regular time intervals. This ensures that no physically-unstable structures can exist in the scene. The interval duration should be set to a value that guarantees at least one execution before every decision-making step of the robot. 

%
%

\subsection{Task strategy planning}
\label{sec:bsp}
As mentioned above, the objective of our task is to minimize the time needed to unload a full truck, by selecting the best action to execute at runtime. This is essentially a sequential decision-making problem under uncertainty since the
robot cannot observe the packet's weight, friction coefficient, etc.

The goal of the strategy-planning module is to find a best action sequence, i.e., a \textit{strategy}, for a given scenario. 
It iteratively evaluates each possible action sequence in simulation and returns the best action sequence that takes the minimal time to unload the target packets. 
We run the strategy planner on many different scenarios, so that we can construct a \textit{strategy   library}.
At runtime, we use the constructed strategy library to choose the strategy that is deemed the most suitable for the current scenario using machine learning techniques by formulating it as a multi-class classification task.

For this strategy planning module, we employed POMHDP---a search-based belief-space planner~\cite{kim2019pomhdp}.
It is a sample-efficient belief-space planner that bootstraps domain-specific knowledge in the form of heuristics. However, it still requires a large number of simulations to evaluate many possible action sequences under motion and sensing  uncertainty. 
It represents the probabilistic distribution of the state as a set of particles (i.e., sampled states), and it iteratively simulates sequential action execution processes in order to capture the system's stochasticity.
Thus, the time complexity of the strategy planning is proportional to
(i)~$T_{\text{sim}}$, the time to execute an action (pick, sweep, etc,) in simulation,
(ii)~$N_{\text{part}}$, the number of particles,
(iii)~$N_{\text{action}}$, the number of possible actions and
(iv)~$N_{\text{iter}}$, the number of iterations used by POMHDP until convergence or timeout.
%
In order to obtain optimal strategies, we need to instantiate a large number of
particles $N_{part}$, evaluate all possible actions, and iterate until
convergence. Hence, any speedups obtained by improving simulation speed is
critical in planning optimal strategies.

\subsection{Example---Adaptive fidelity within strategy planning}
\begin{figure}[t]
  \centering
  \includegraphics[height=5.cm]{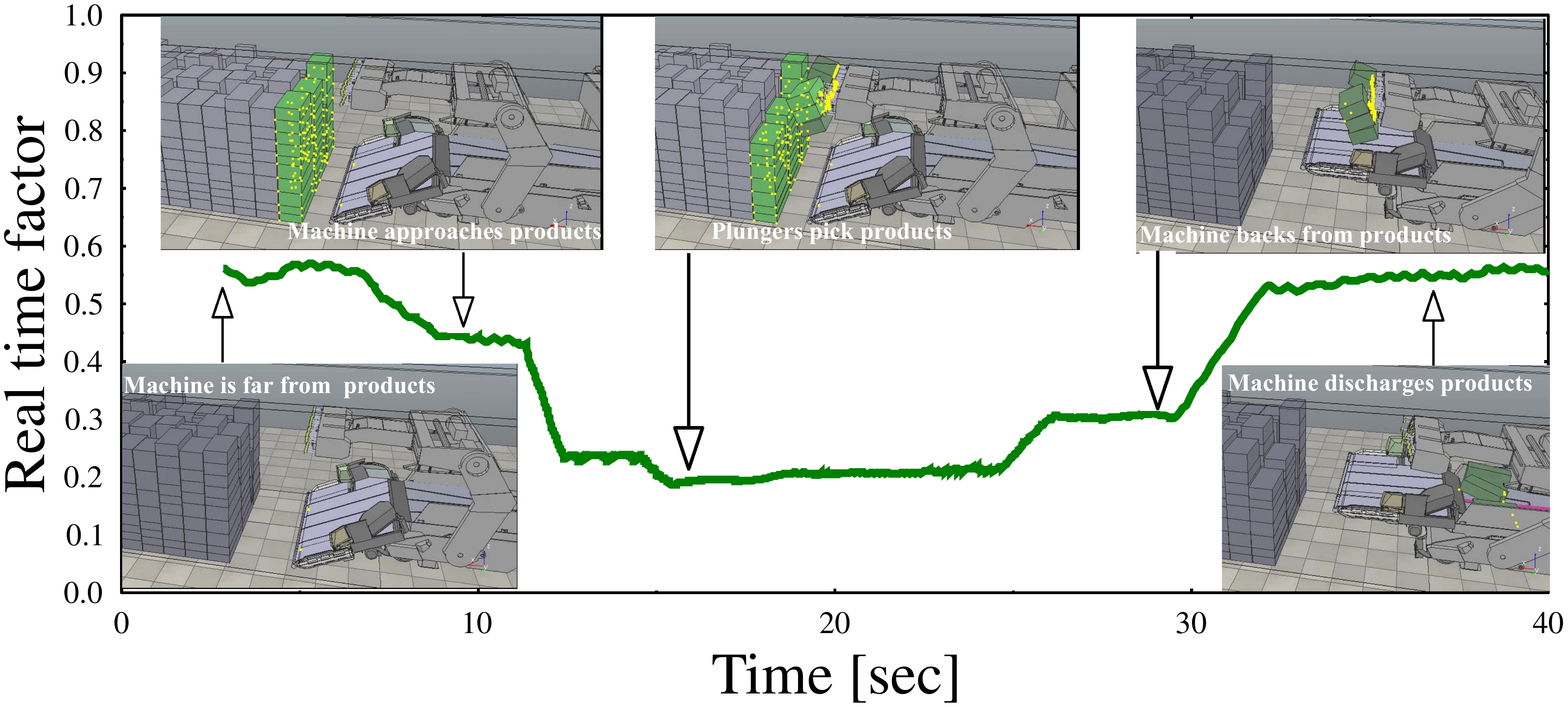}
 \caption{Visualization of how the speed of simulation changes with time using our approach for adaptive fidelity. Real time factor denotes the number of seconds elapsed in simulation time in one second of real time (a higher real time factor implies faster simulation.)}
  \label{fig:adaptive}
\end{figure}

To demonstrate how our framework performs, consider the action of picking packets considered by the planning algorithm.
Here, the robot starts at an idle state at a distance from the  packets. 
Consequently, all packets are in low-fidelity and the simulation runs at a high speed.
As the robot approaches the packets in order to perform a pick operation, the fidelity of packets that may be in contact with the robot (namely, those that intersect the robot's bounding box) are toggled into high-fidelity mode.
As the robot further approaches the packets, the plungers at the end of its arm are extended and their respective scripts are toggled into high-fidelity mode.
Here the simulation speed is the lowest due to the complex interactions between the boxes and the robot plungers.
Finally, after the packets are picked, the robot retracts its plungers and moves away from the packets (in order to allow its sensors to view the entire scene), all boxes are toggled back to low-fidelity and simulation speed increases.
For a visualization, see Fig.~\ref{fig:adaptive} which plots the (instantaneous) real-time factor, i.e., the ratio of the simulated time to the real-time needed for that simulation, as a function of simulation time.  

\section{Results} \label{sec:results}

For our experiments, we compare the accuracy and speed of our adaptive-fidelity mode with the high-fidelity mode. This method of reporting accuracy is borrowed from work by Erez et al.~\cite{erez2015simulation}.
%
In their work, while presenting results for large-scale simulation, Erez et al. state that 
\begin{quote}
  ``\emph{one can imagine future applications
involving \dots\, robots manipulating many movable objects \dots\, scaling to
larger systems becomes essential}''.
\end{quote}
\noindent
That future is here, and our simulation needs to scale with the many packets that need to be unloaded.

\begin{figure}[t]
  \centering
  \begin{subfigure}{0.32\textwidth}
    \includegraphics[clip,trim=12cm 3cm 0cm 3cm, width=0.9\textwidth]{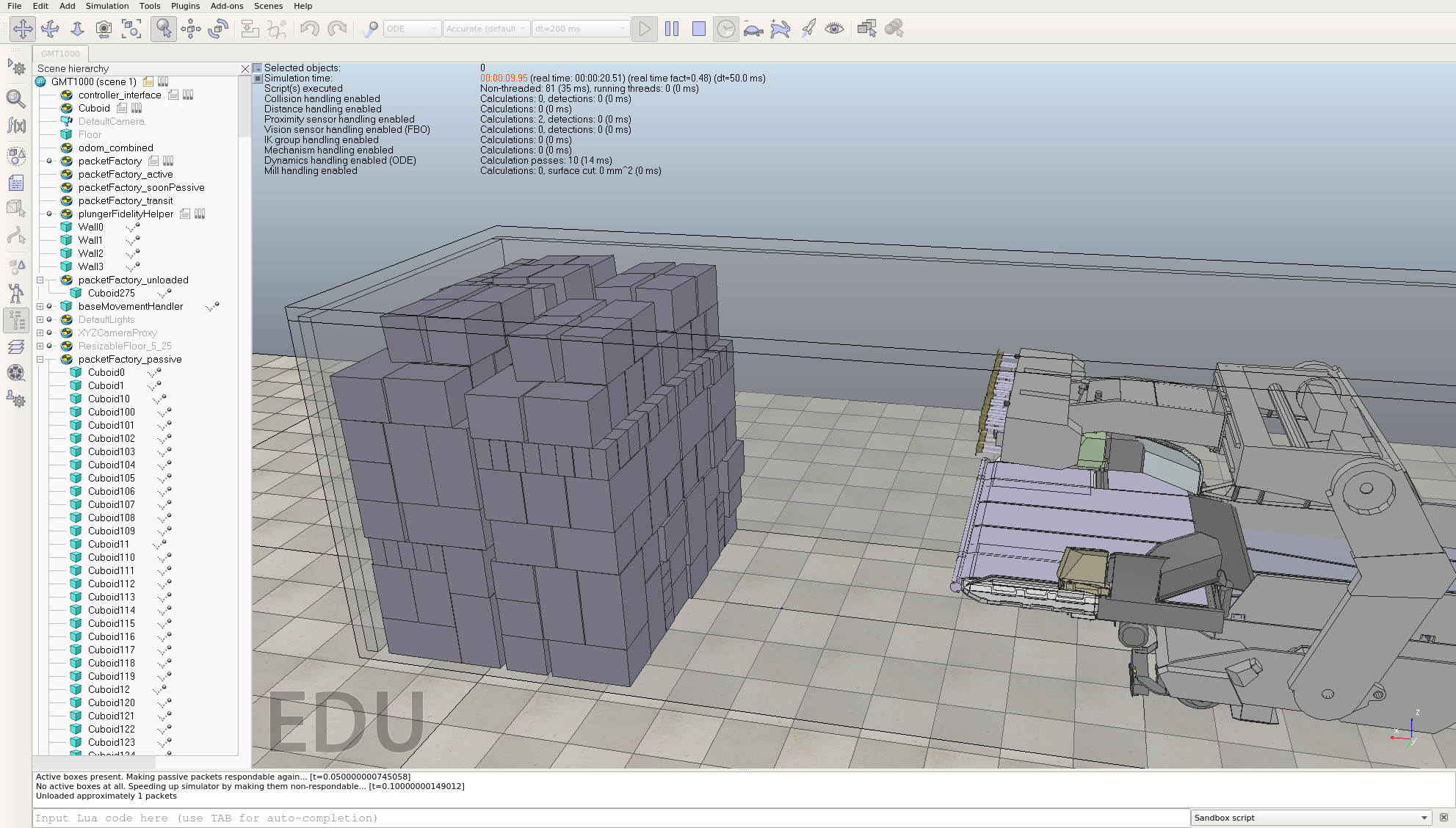}
    \caption{Scenario A.}
    \label{fig:view4}
  \end{subfigure}
  \begin{subfigure}{0.32\textwidth}
    \includegraphics[clip,trim=12cm 3cm 0cm 3cm, width=0.9\textwidth]{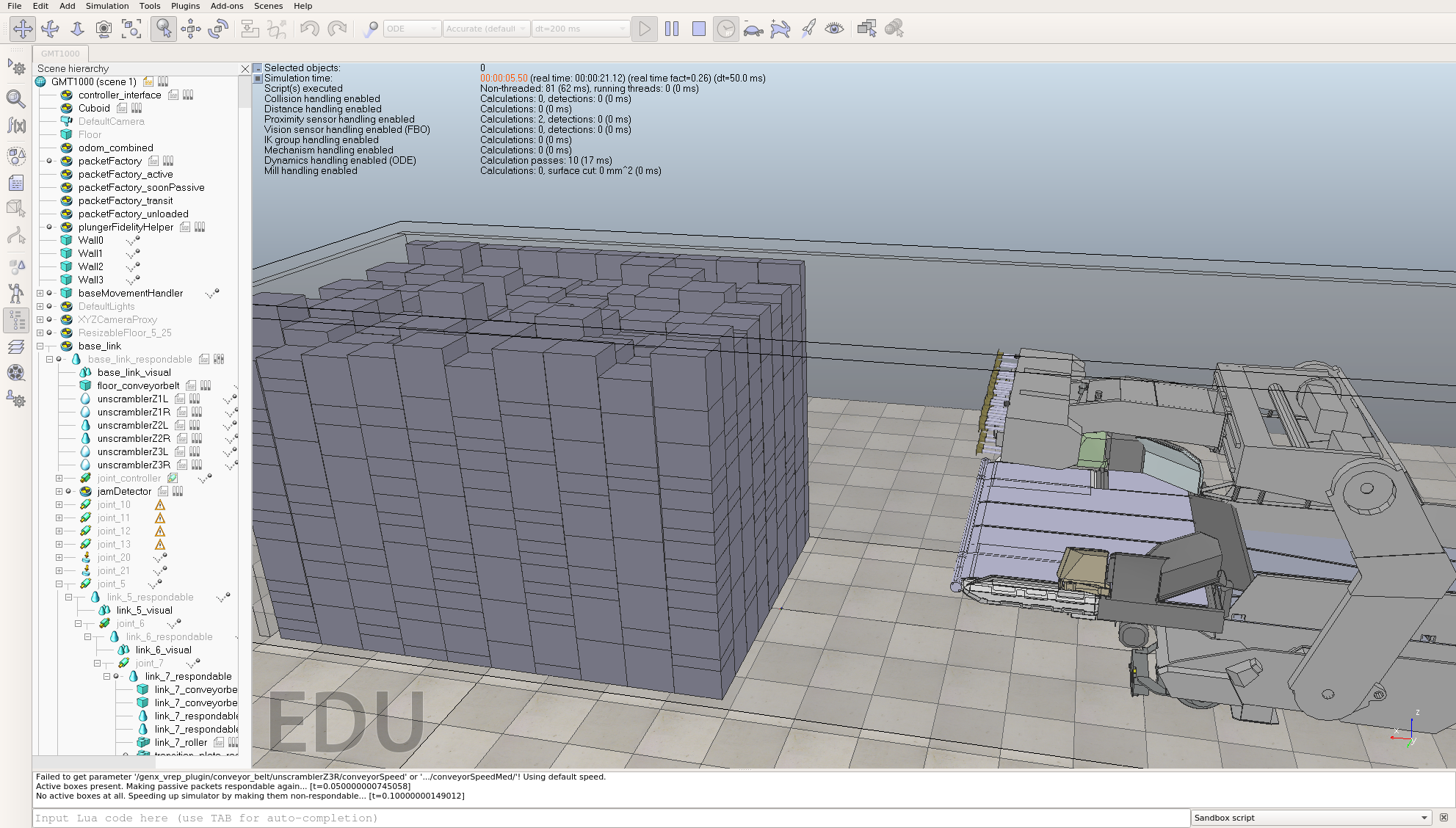}
    \caption{Scenario B.}
    \label{fig:view6}
  \end{subfigure}
  \begin{subfigure}{0.32\textwidth}
    \includegraphics[clip,trim=12cm 3cm 0cm 3cm, width=0.9\textwidth]{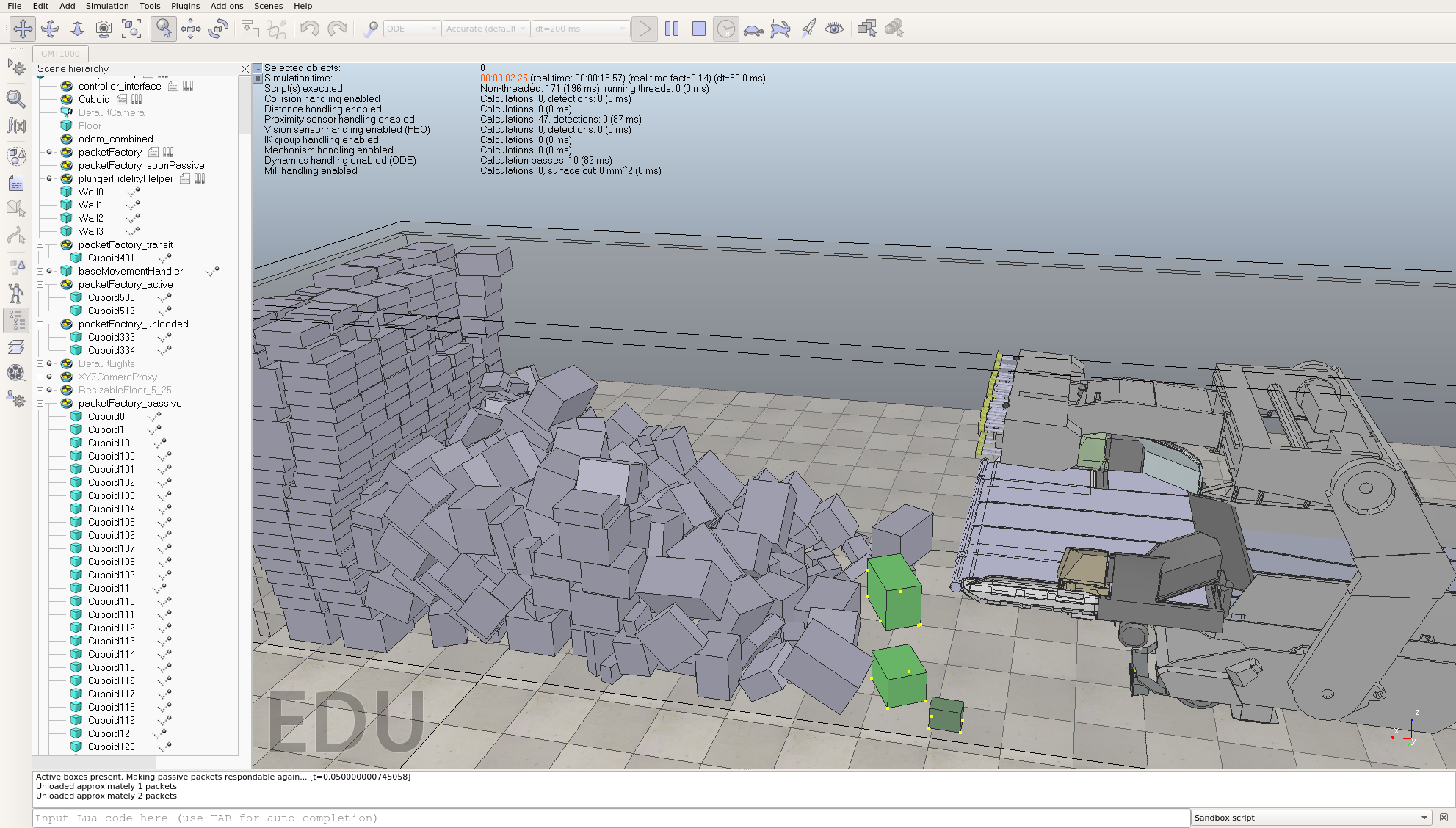}
    \caption{Scenario C.}
    \label{fig:view9A}
  \end{subfigure}
  \caption{Screenshots of three representative scenarios of truck unloading.}
  \label{fig:views}
\end{figure}

\subsection{Experimental setup} \label{ssec:description-of-experiments}

Our task-informed fidelity manager for the plungers on the robot arm, described in Sec.~\ref{ssec:simulation-fidelity-management}, leverages the fact that we know in advance when the plungers are being used. Thus, we can reliably toggle between high-fidelity and low-fidelity states of the plungers without affecting the accuracy of the simulation. 

However, our fidelity manager for the packets in the truck can possibly alter the simulation results. When using this fidelity manager, packets far away from the robot will not be simulated in high-fidelity mode, based on the assumption that the robot cannot directly interact with those objects. They will thus remain static. Without using the fidelity manager, on the other hand, all packets will continuously be simulated in high-fidelity mode. When the packets are stacked tightly against each other, a force applied to the front of the pile might propagate to the back of the pile and cause small displacements. These displacements would not occur when using the adaptive fidelity manager.

In our experiments, we tested multiple versions of our adaptive fidelity manager. Each version uses a different size for the inflated robot bounding box. By increasing the bounding box volume, more packets will be simulated in high-fidelity, which should increase the accuracy of the simulation. We created three adaptive-fidelity managers ``AdFi-1'', ``AdFi-2'', ``AdFi-3'', which use a robot bounding box radius that is inflated by 1.00, 0.50, and 0.01 meters, respectively.

\begin{figure}
  \centering
  \includegraphics[clip,trim=0cm 0cm 0cm 0cm, width=0.6\textwidth,height=3.5cm]{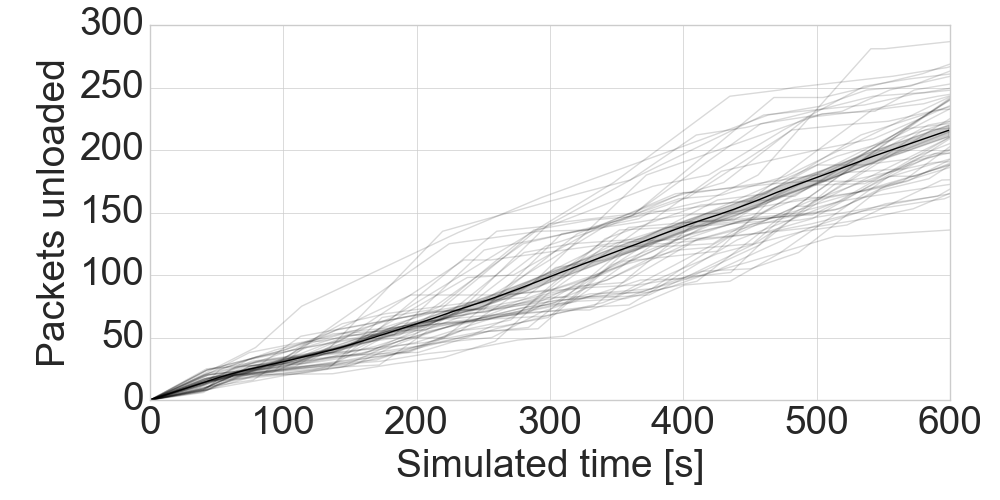}
  \caption{Number of packets unloaded when simulating scenario~A for 50 times in HiFi mode, depicting the large stochasticity in our system.}
  \label{fig:rates}
\end{figure}

We used three typical truck scenarios, which are shown in Fig.~\ref{fig:views}. Scenario A,~B, and~C contain 491, 1007, and 539 packets, respectively. Running such large-scale simulation scenes in V-REP with the physics engine ODE manifests a high degree of stochasticity. For this reason, we repeated each experiment 50 times and averaged the obtained unloading rates. 
Fig.~\ref{fig:rates} illustrates the scale of the variance we observed between individual simulations for ten minutes of truck unloading, even when using high fidelity.

\subsection{Speed versus accuracy} \label{ssec:speed_and_accuracy}

We repeatedly simulated ten minutes of the same truck-unloading task using the above adaptive-fidelity managers and compared the results to our base line, the ``HiFi'' mode, which consists of the same simulation without using any fidelity managers. Given that the objective of the robot is to maximize the packet unloading rate, we used this rate as the metric for our accuracy comparison. An ideal implementation of the adaptive-fidelity manager should therefore speed up the simulation without noticeably altering the robot's unloading performance. 

\begin{figure}[tb]
  \begin{subfigure}{0.325\textwidth}
    \includegraphics[clip,trim=0cm 0cm 0.0cm 0.0cm, width=0.95\textwidth,height=3.7cm]{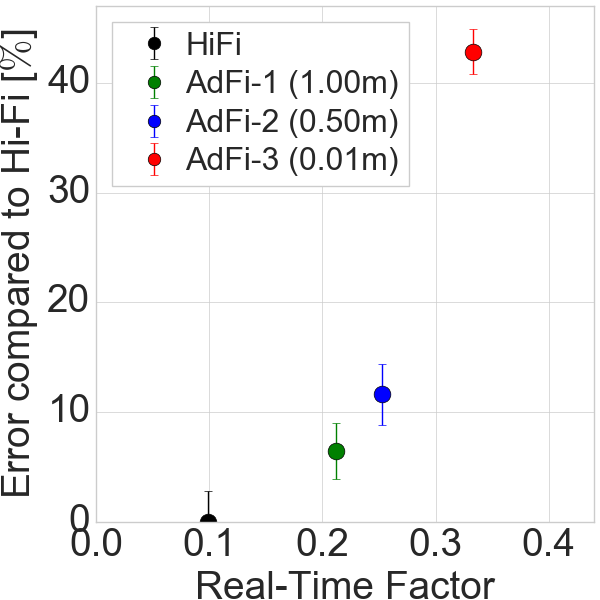}
    \caption{Scenario A (HiFi unload rate: 1295 packets/hr).}
    \label{fig:comparisonA}
  \end{subfigure}
  \begin{subfigure}{0.325\textwidth}
    \includegraphics[clip,trim=0cm 0cm 0.0cm 0.0cm, width=0.95\textwidth,height=3.7cm]{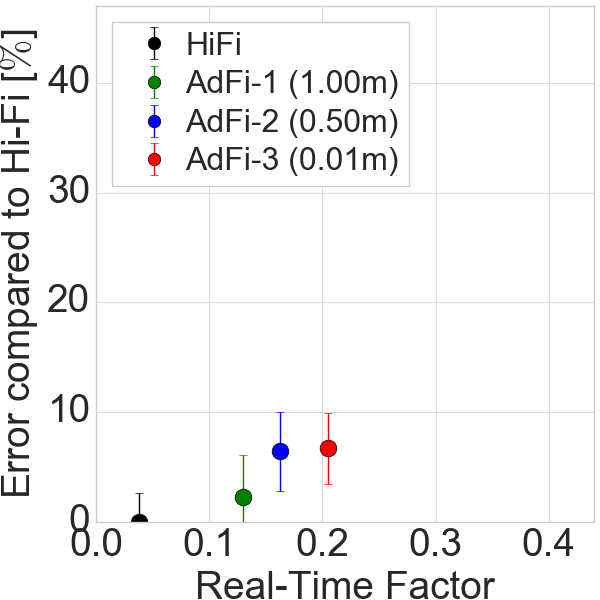}
    \caption{Scenario B (HiFi unload rate: 938 packets/hr).}
    \label{fig:comparisonB}
  \end{subfigure}
  \begin{subfigure}{0.325\textwidth}
    \includegraphics[clip,trim=0cm 0cm 0.0cm 0.0cm, width=0.95\textwidth,height=3.7cm]{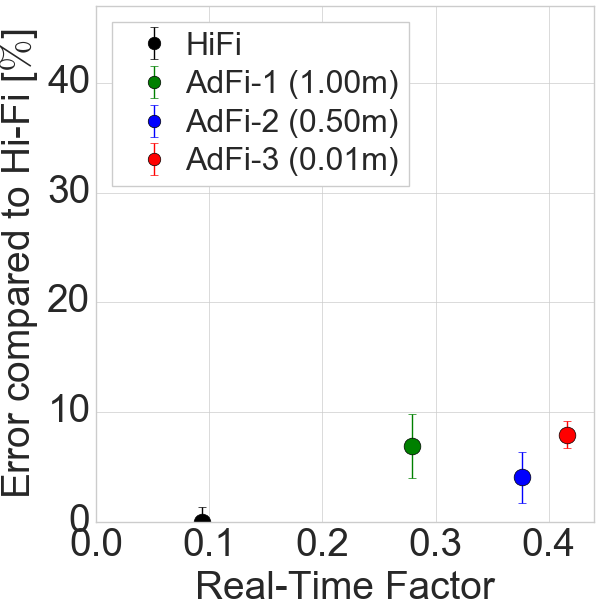}
    \caption{Scenario C (HiFi unload rate: 1583 packets/hr).}
    \label{fig:comparisonC}
  \end{subfigure}
  \caption{Error of the unloading rate for each fidelity mode when compared to the simulations in high fidelity, shown as the absolute value of the percent error.}
  \label{fig:comparison}
\end{figure}

Fig.~\ref{fig:comparison} summarizes the results of our experiments for each adaptive-fidelity mode. 
Specifically, we plot the error as a function of the \emph{real-time factor}, which is the ratio of simulation time and computation time.
A higher real-time factor thus corresponds to a faster simulation.

As expected, all variants of adaptive-fidelity mode result in a faster simulation, when compared to high fidelity with a direct correlation between the size of the inflated robot bounding box and simulation speed. 
Specifically, a smaller bounding box reduces the number of packets simulated in high-fidelity and therefore speeds up the simulation.

Concerning accuracy, the overall trend visible in Fig.~\ref{fig:comparison} is that faster fidelity modes deviate more from the base line. 
A possible explanation is that a smaller robot bounding box results in a lower number of packets being simulated in high-fidelity. %
This is clearly visible for scenario~A. 
In the other two tested scenarios, this trend is less obvious. 
Here, all modes display an error of less than 10\%. This implies that for scenario~B and~C a very small bounding box inflation is sufficient to obtain accurate results. 
As discussed in Sec.~\ref{ssec:description-of-experiments}, the dimensions of the dynamic shapes as well as their structural placement in the scene can strongly influence how much bounding-box inflation is required to ensure accurate results. 
Finally, we note that the results indicate that the speed improvement is more significant for scenarios where the majority of packets are located far away from the robot, such as scenarios~B and~C.

\subsection{Effective planning versus simulation fidelity}
\label{sec:bsp_results}

\begin{figure}[t]
  \centering
  \begin{subfigure}{0.475\textwidth}
    \includegraphics[height=5.94cm]{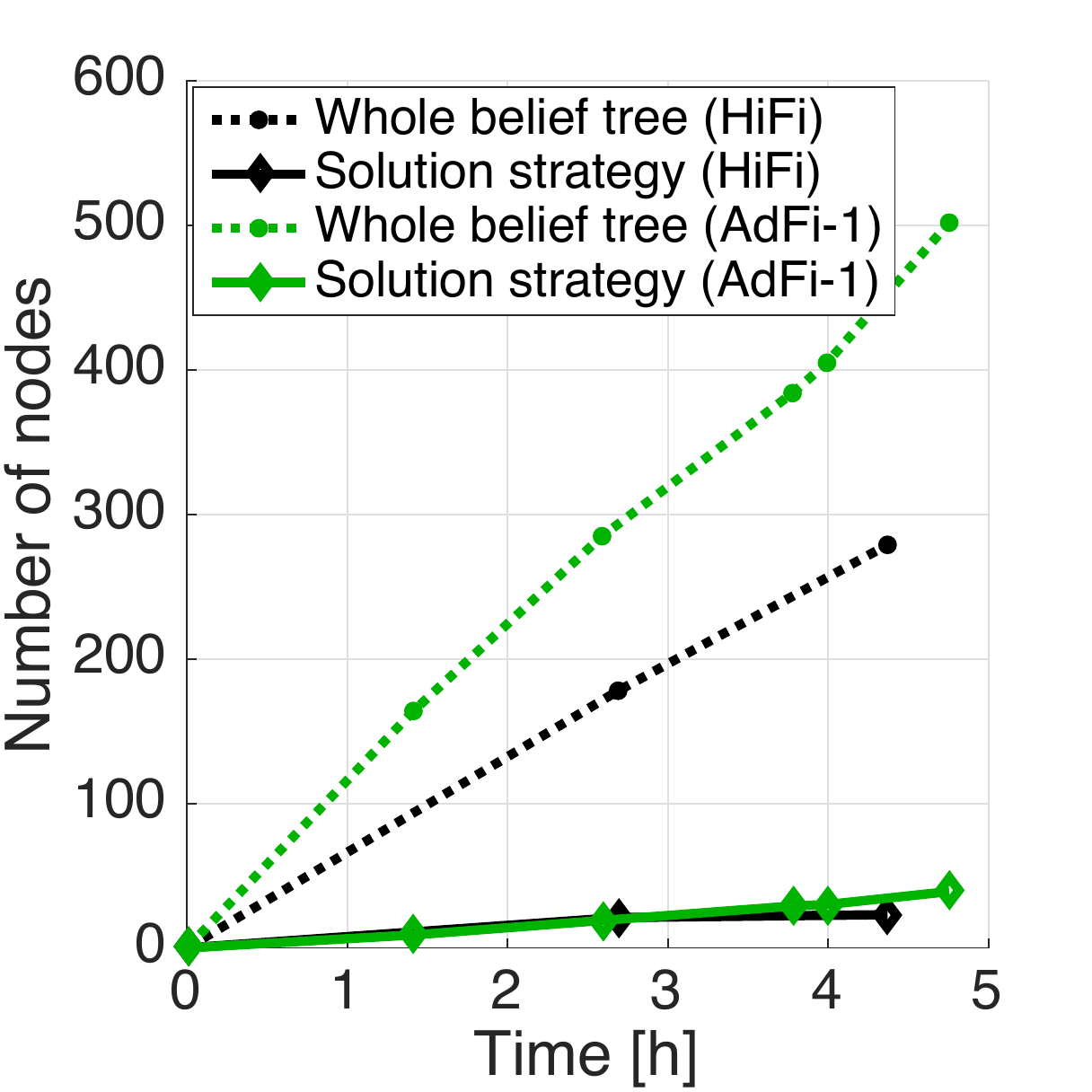}
    \caption{Planning efficiency.}
    \label{fig:bsp_nodes}
  \end{subfigure}
  \begin{subfigure}{0.475\textwidth}
    \includegraphics[clip,trim=0cm 0.65cm 0cm 0.0cm, height=5.94cm]{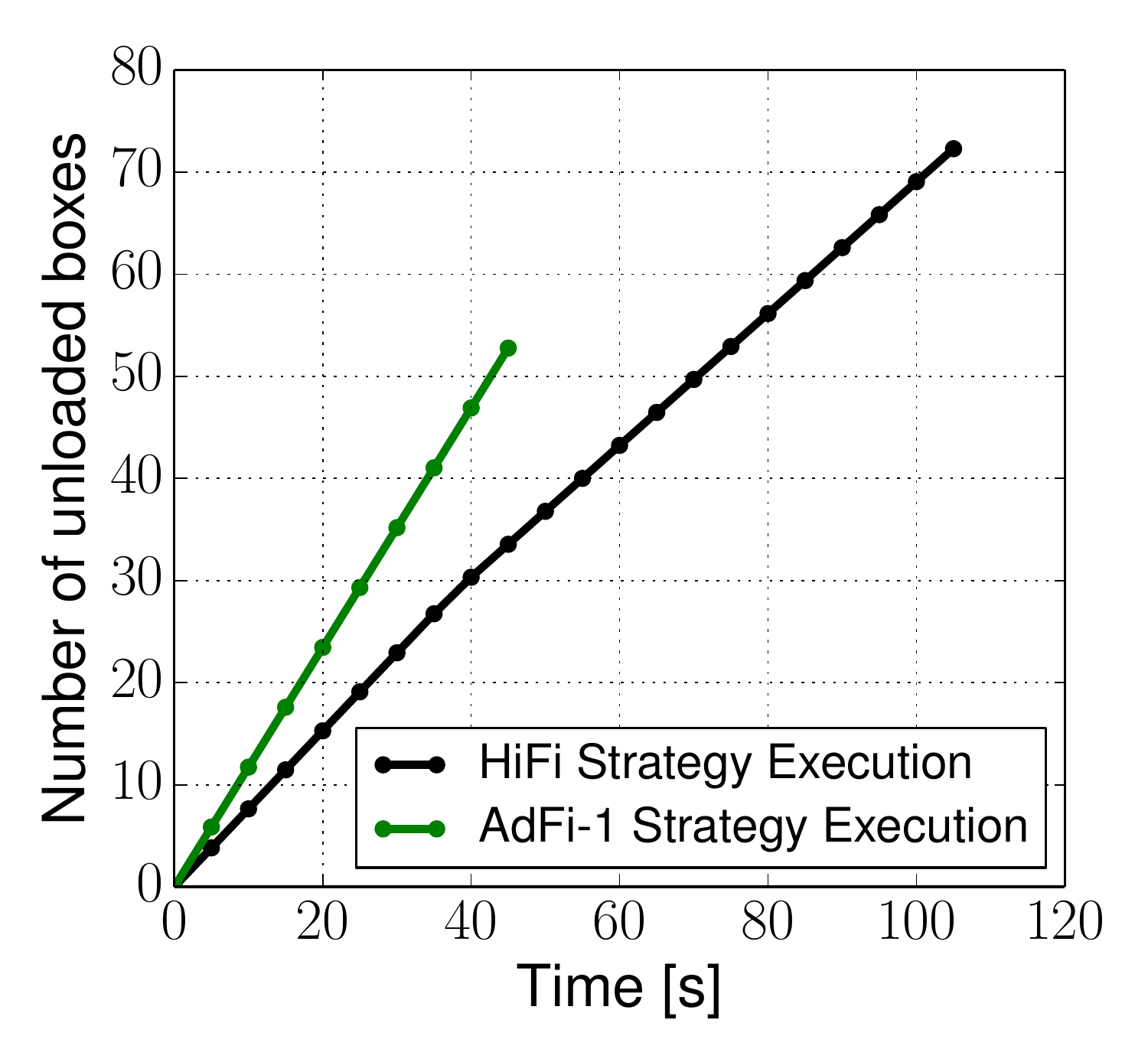}
    \caption{Planned strategy evaluation.}
    \label{fig:bsp_eval}
  \end{subfigure}
  \caption{Efficiency of strategy planning efficiency in HiFi and AdFi-1 modes (left) and evaluation of the strategies generated in each mode within 5 hours (right) for Scenario~A.
  }
  \label{fig:bsp_results}
\end{figure}

In  the previous section, we demonstrated that our approach for adaptive fidelity allows for a significant speedup in simulation with relatively little error on the quality of the simulation.
A natural question to ask is should one make this compromise?
In this section we demonstrate empirically that the answer to this question is a clear ``yes''.

Recall (Sec.~\ref{sec:bsp}) that we use POMHDP to generate strategies in an offline phase. These strategies are then used in an online phase to unload the truck trailers.
We ran the simulation both in HiFi and in AdFi-1 for five hours in order to generate strategies using POMHDP.
We then compared the effectiveness of said strategies in the online phase where the simulator was run in high-fidelity mode.

Running in  AdFi-1 allowed POMHDP to search a larger portion of the belief space when compared to HiFi.
This is demonstrated by the number of belief-space nodes that POMHDP traversed (Fig.~\ref{fig:bsp_nodes}).
This immediately corresponds to better strategies as can be seen in Fig.~\ref{fig:bsp_eval}.
Roughly speaking, the strategy generated using AdFi-1 unloads boxes $1.5\times$ faster than the strategy generated using HiFi.

\subsection{Application to machine learning algorithms}
\label{sec:train-optim-polic}
\begin{table}[t]
   \centering
   \begin{tabular}{|c|c|c|c|}
     \hline
     \textbf{Fidelity setting}& \textbf{Training accuracy}&
 \textbf{Test accuracy} &
\textbf{Training time [hours]}
     \\
     \hline
     \textbf{AdFi-1} & 0.82 & 0.68 & 1\\
     \hline
     \textbf{HiFi} & 0.79 & 0.59 & 1\\
     \hline
     \textbf{HiFi} & 0.84 & 0.7 & 2\\
     \hline
   \end{tabular}

   \caption{Training and test statistics for a classifier that
     chooses between a pick and a sweep action given the current state
     of the environment}
   \label{tab:training}

 \end{table}
In our final experiment, we demonstrate the potential of our approach
in generating data for machine learning algorithms.
We trained a binary logistic regression classifier to choose between pick and sweep action in both AdFi and HiFi settings for Scenario~A. 
For each step, we run both pick and sweep actions independently (by
resetting the simulator after the first action) and label the current
state with the action that resulted in the largest number of boxes
unloaded. This constitutes our training dataset.
%
%
We then test the accuracy of our classifier on a test set obtained using a variant of Scenario~A.
Table~\ref{tab:training} shows that AdFi-1 and HiFi obtain comparable
test accuracies where AdFi-1 required half the time to train the
classifier, when compared to HiFi. Given the same training time of 1 hour,
AdFi-1 achieves a better test accuracy than HiFi owing to better real
time factor which leads to more data being collected.
This shows the efficacy of using the proposed
task-informed adaptive fidelity settings that allow faster training.

\section{Conclusion and future work} \label{sec:conclusion}

We presented a principled approach for speeding up robot simulators by actively modifying the dynamic properties of certain objects in the simulation scene. As a use case, we applied this framework to a real robot application, unloading packets from a truck onto a warehouse conveyor belt. Our framework includes an adaptive fidelity manager that toggles the dynamic state of the robot end-effector without affecting the fidelity of the simulation. Another fidelity manager was developed, which controls the dynamic properties of packets in the scene, based on whether they are relevant to the defined robot task. We show that the latter module can be tuned to achieve a speed-up of the simulator while only minimally altering its outcome. Consequently, our architecture can be used as a tool to accelerate the training of robot policies that depend on the outputs of complex and slow simulations.
As future work, we wish to test our approach on different robot applications, and investigate how to set the parameters used by the adaptive fidelity manager to achieve large speedups without sacrificing simulation accuracy.

\bibliography{references}  

\end{document}
